\documentclass[runningheads]{llncs}
\usepackage{graphicx}
\usepackage{todonotes}
\usepackage{url}
\usepackage{cite}
\usepackage{bbm}
\usepackage{amsmath}
\usepackage{amssymb}
\usepackage{multirow}
\usepackage{algorithm,algorithmic}
\usepackage{comment}
\usepackage{graphicx}
\usepackage{bbm}
\usepackage{bm}
\usepackage{dsfont}
\usepackage{mathtools}
\usepackage{adjustbox}
\usepackage{hyperref}

\begin{document}
\title{vMFNet: Compositionality Meets Domain-generalised Segmentation}
\titlerunning{vMFNet: Compositionality Meets Domain-generalised Segmentation}

\author{Xiao Liu\inst{1,4}, Spyridon Thermos\inst{2}, Pedro Sanchez\inst{1,4}, Alison Q. O'Neil\inst{1,4} \and Sotirios A. Tsaftaris\inst{1,3,4}} 
%index{Liu, Xiao}
%index{Thermos, Spyridon}
%index{Sanchez, Pedro}
%index{O'Neil, Alison Q.}
%index{Tsaftaris, Sotirios A.}
\authorrunning{X. Liu et al.}
\institute{School of Engineering, University of Edinburgh, Edinburgh EH9 3FB, UK 
\and
AC Codewheel Ltd
\and
The Alan Turing Institute, London, UK \and
Canon Medical Research Europe Ltd., Edinburgh, UK\\
\email{Xiao.Liu@ed.ac.uk}}

\maketitle              % typeset the header of the contribution
\sloppy
\setcounter{footnote}{0} 
\begin{abstract}
Training medical image segmentation models usually requires a large amount of labeled data. By contrast, humans can quickly learn to accurately recognise anatomy of interest from medical (e.g. MRI and CT) images with some limited guidance. Such recognition ability can easily generalise to new images from different clinical centres. This rapid and generalisable learning ability is mostly due to the compositional structure of image patterns in the human brain, which is less incorporated in medical image segmentation. In this paper, we model the compositional components (i.e. patterns) of human anatomy as learnable von-Mises-Fisher (vMF) kernels, which are robust to images collected from different domains (e.g. clinical centres). The image features can be decomposed to (or composed by) the components with the composing operations, i.e. the vMF likelihoods. The vMF likelihoods tell how likely each anatomical part is at each position of the image. Hence, the segmentation mask can be predicted based on the vMF likelihoods. Moreover, with a reconstruction module, unlabeled data can also be used to learn the vMF kernels and likelihoods by recombining them to reconstruct the input image. Extensive experiments show that the proposed vMFNet achieves improved generalisation performance on two benchmarks, especially  when annotations are limited. Code is publicly available at: \url{https://github.com/vios-s/vMFNet}.

\keywords{Compositionality \and Domain generalisation \and Semi-supervised learning \and Test-time training \and Medical image segmentation.}
\end{abstract}

\section{Introduction}
Deep learning approaches can achieve impressive performance on medical image segmentation when provided with a large amount of labeled training data \cite{bernard2018deep, chen2020frontiers}. However, shifts in data statistics, i.e. \textit{domain shifts} \cite{zhang2020generalising, zakazov2021anatomy}, can heavily degrade deep model performance on unseen data \cite{prados2017spinal, mnms, li2021atrialgeneral}. By contrast, humans can learn quickly with limited supervision and are less likely to be affected by such domain shifts, achieving accurate recognition on new images from different clinical centres. Recent studies \cite{tokmakov2019learning, liu2021learning} argue that this rapid and generalisable learning ability is mostly due to the compositional structure of concept components or image patterns in the human brain. For example, a clinical expert usually remembers the patterns (components) of human anatomy after seeing many medical images. When searching for the anatomy of interest in new images, patterns or combinations of patterns are used to find where and what the anatomy is at each position of the image. This process can be modeled as learning compositional components (\textit{learning}), composing the components (\textit{composing}), and performing downstream tasks (\textit{performing}). This compositionality has been shown to improve the robustness and explainability of many computer vision tasks \cite{tokmakov2019learning, huynh2020compositional, kortylewski2020compositional} but is less studied in medical image segmentation.
\begin{figure}[t]
\includegraphics[width=\textwidth]{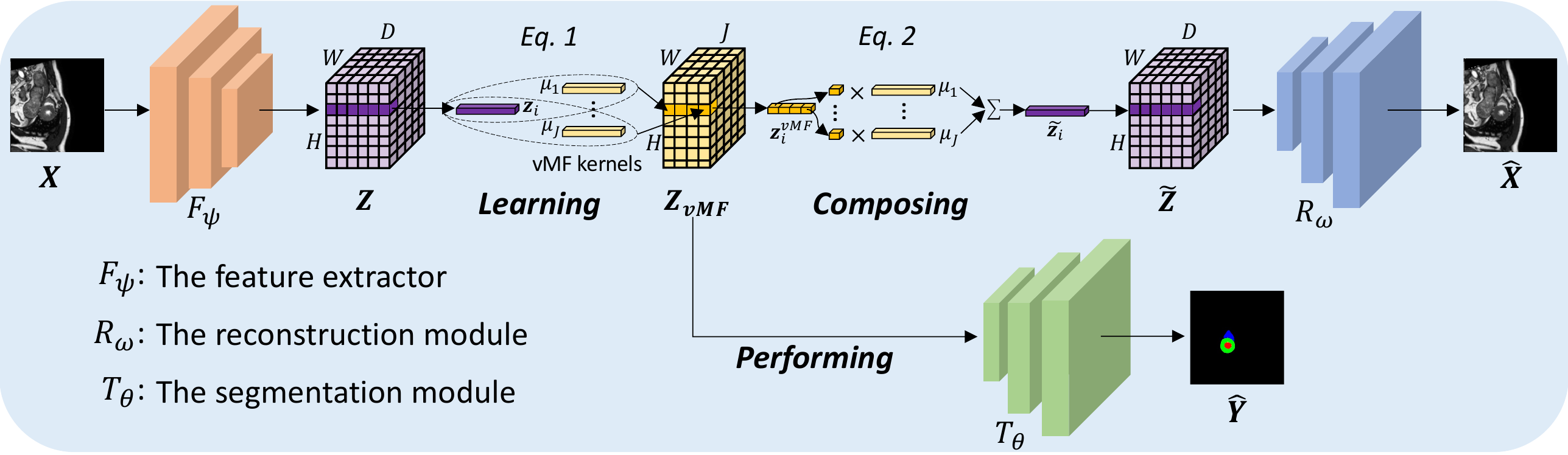}
\centering
\caption{The overall model design of vMFNet. The notations are specified in Section~\ref{sec:method}.} \label{fig::model}
\end{figure}

In this paper, we explore compositionality in domain-generalised medical image segmentation and propose vMFNet. The overall model design is shown in Fig.~\ref{fig::model}. Motivated by Compositional Networks \cite{kortylewski2020compositional} and considering medical images are first encoded by deep models as deep features, we model the compositional components of human anatomy as learnable von-Mises-Fisher (vMF) kernels (\textit{learning}).\footnote{vMF kernels are similar to prototypes in  \cite{snell2017prototypical}. However, prototypes are usually calculated as the mean of the feature vectors for each class using the ground-truth masks. vMF kernels are learnt as the cluster centres of the feature vectors.} The features can be projected as vMF likelihoods that define how much each kernel is activated at each position. We then compose the kernels with the normalised vMF likelihoods to reconstruct the input image (\textit{composing}). Hence, the kernels and the corresponding vMF likelihoods can also be learnt end-to-end with unlabeled data. In terms of a downstream segmentation task, the spatial vMF likelihoods are used as the input to a segmentation module to predict the segmentation mask (\textit{performing}). In this case, the prediction is based on the activation of the components at each position of the image.  

For data from different domains, the features of the same anatomical part of images from these domains will activate the same kernels. In other words, the compositional components (kernels) are learnt to be robust to the domain shifts. To illustrate such robustness, we consider the setting of domain generalisation \cite{li2018learning}. With data from multiple source domains available, domain generalisation considers a strict but more clinically applicable setting in which no information about the target domain is known during training. We also consider test-time training \cite{hu2021fully, valvano2021stop, karani2021test, he2021autoencoder} with our model on unseen test data at inference time, i.e.\ test-time domain generalisation \cite{iwasawa2021test}. We observe that by freezing the kernels and the segmentation module and fine-tuning the rest of the model to better reconstruct the test data, improved generalisation performance is obtained. This is due to the fact that medical images from different domains are not significantly different; new components do not need to be learned, rather the composing operations need to be fine-tuned.

\vspace{5pt}\noindent\textbf{Contributions}:
\begin{itemize}
    \item Inspired by the human recognition process, we design a semi-supervised compositional model for domain-generalised medical image segmentation.
    \item  We propose a reconstruction module to compose the vMF kernels with the vMF likelihoods to facilitate reconstruction of the input image, which allows the model to be trained also with unlabeled data.
    \item We apply the proposed method to two settings: semi-supervised domain generalisation and test-time domain generalisation.  
    \item Extensive experiments on cardiac and gray matter datasets show improved performance over several baselines, especially when annotations are limited.
\end{itemize}

\section{Related work}
\textbf{Compositionality}: Compositionality has been mostly incorporated for robust image classification \cite{tokmakov2019learning, kortylewski2020compositional} and recently for compositional image synthesis \cite{liu2021learning, arad2021compositional}. Among these work, Compositional Networks \cite{kortylewski2020compositional} originally designed for robust classification under object occlusion is easier to extend to pixel-wise tasks as it learns spatial and interpretable vMF likelihoods. Previous work integrates the vMF kernels and likelihoods \cite{kortylewski2020compositional} for object localisation \cite{yuan2021robust} and recently for nuclei segmentation (with the bounding box as supervision) in a weakly supervised manner \cite{zhang2021light}. We rather use the vMF kernels and likelihoods for domain-generalised medical image segmentation and learn vMF kernels and likelihoods with unlabeled data in a semi-supervised manner.

\noindent \textbf{Domain generalisation}: Several active research directions handle this problem by either augmenting the source domain data \cite{zhang2020generalising, chen2021cooperative}, regularising the feature space \cite{carlucci2019domain, huang2021fsdr}, aligning the source domain features or output distributions \cite{li2020domain}, carefully designing robust network modules \cite{gu2021domain}, or using meta-learning to approximate the possible domain shifts across source and target domains \cite{dou2019domain, liu2020shape, liu2021feddg, liu2021semi}. Most of these methods consider a fully supervised setting. Recently, Liu et al.\ \cite{liu2021semi} proposed a gradient-based meta-learning model to facilitate semi-supervised domain generalisation by integrating disentanglement. Extensive results on a common backbone showed the benefits of meta-learning and disentanglement. Following \cite{liu2021semi}, Yao et al.\ \cite{yao2022enhancing} adopted a pre-trained ResNet \cite{he2016deep} as a backbone feature extractor and augmented the source data by mixing MRI images in the Fourier domain and employed pseudo-labelling to leverage the unlabelled data. Our proposed vMFNet aligns the image features to the same vMF distributions to handle the domain shifts. The reconstruction further allows the model to handle semi-supervised domain generalisation tasks.

\section{Proposed method}
\label{sec:method}
We propose vMFNet, a model consisting of three modules; the feature extractor $\bm{F}_\psi$, the task network $\bm{T}_\theta$, and the reconstruction network $\bm{R}_\omega$, where $\psi$, $\theta$ and $\omega$ denote the network parameters. Overall, the compositional components are learned as vMF kernels by decomposing the features extracted by $\bm{F}_\psi$. Then, we compose the vMF kernels to reconstruct the image with $\bm{R}_\omega$ by using the vMF likelihoods as the composing operations. Finally, the vMF likelihoods that contain spatial information are used to predict the segmentation mask with $\bm{T}_\theta$. The decomposing and composing are shown in Fig. \ref{fig::model} and detailed below.

\subsection{Learning compositional components}

% Representations that are generalisable incorporate knowledge about the world structure [cite causal or bengio papers or disentanglement tutorial] that generated the data. Medical images have an inherent compositional structure. For example, the cardiac system is composed by many organs along with vessels, these organs are themselves composed of other substructures. In this paper, we aim to capture the natural compositional nature of the human body in order to improve generalisation.

The primal goal of learning compositional components is to represent deep features in a compact low dimensional space. We denote the features extracted by $\bm{F}_\psi$ as $\mathbf{Z} \in \mathbb{R}^{H\times W\times D}$, where $H$ and $W$ are the spatial dimensions and $D$ is the number of channels. The feature vector $\mathbf{z}_i\in \mathbb{R}^{D}$ is defined as a vector across channels at position $i$ on the 2D lattice of the feature map. We follow Compositional Networks \cite{kortylewski2020compositional} to model $\mathbf{Z}$ with $J$ vMF distributions, where the learnable mean of each distribution is defined as vMF kernel $\boldsymbol \mu_{j} \in \mathbb{R}^{D}$. To allow the modeling to be tractable, the variance $\sigma$ of all distributions are fixed. In particular, the vMF likelihood for the $j^{th}$ distribution at each position $i$ can be calculated as: 
\begin{equation}
    p(\mathbf{z}_i|\boldsymbol \mu_{j}) = {C(\sigma)}^{-1} \cdot {e^{\sigma \boldsymbol \mu_{j}^{T} \mathbf{z}_i}} , \text{ s.t. } ||\boldsymbol \mu_{j}||=1, 
\label{eq::vmf}
\end{equation}
where $||\mathbf{z}_i||=1$ and $C(\sigma)$ is a constant. After modeling the image features with $J$ vMF distributions with Eq.\ \ref{eq::vmf}, the vMF likelihoods $\mathbf{Z}_{vMF} \in \mathbb{R}^{H\times W\times J}$ can be obtained, which indicates how much each kernel is activated at each position. Here, the vMF loss $\mathcal{L}_{vMF}$ that forces the kernels to be the cluster centres of the features vectors is defined in \cite{kortylewski2020compositional} as $
    \mathcal{L}_{vMF}(\boldsymbol \mu, \mathbf{Z}) = -{(HW)}^{-1}\sum_{i} \operatorname*{max}_{j} \boldsymbol \mu_{j}^T \mathbf{z}_i.
$
This avoids an iterative EM-type learning procedure. Overall, the feature vectors of different images corresponding to the same anatomical part will be clustered and activate the same kernels. In other words, the vMF kernels are learnt as the components or patterns of the anatomical parts. Hence, the vMF likelihoods $\mathbf{Z}_{vMF}$ for the features of different images will be aligned to follow the same distributions (with the same means). In this case, the vMF likelihoods can be considered as the content representation in the context of content-style disentanglement \cite{chartsias2019disentangled, liu2020measuring}, where the vMF kernels contain the style information.

\subsection{Composing components for reconstruction}
After decomposing the image features with the vMF kernels and the likelihoods, we re-compose to reconstruct the input image. Reconstruction requires that complete information about the input image is captured \cite{achille2018emergence}. However, the vMF likelihoods contain only spatial information as observed in \cite{kortylewski2020compositional}, while the non-spatial information is compressed as varying kernels $\boldsymbol \mu_{j}, j\in \{1\cdots J\}$, where the compression is not invertible. Consider that the vMF likelihood $p(\mathbf{z}_i|\boldsymbol \mu_{j})$ denotes how much the kernel $\boldsymbol \mu_{j}$ is activated by the feature vector $\mathbf{z}_i$. We propose to construct a new feature space $\mathbf{\widetilde{Z}}$ with the vMF likelihoods and kernels. Let $\mathbf{z}^{vMF}_i\in \mathbb{R}^{J}$ be a normalised vector across $\mathbf{Z}_{vMF}$ channels at position $i$. We devise the new feature vector $\mathbf{\widetilde{z}}_i$ as the combination of the kernels with the normalised vMF likelihoods as the combination coefficients, namely:
\begin{equation}
    \mathbf{\widetilde{z}}_i = \sum_{j=1}^{J} \mathbf{z}^{vMF}_{i, j}  \boldsymbol \mu_{j}, \text{ where } ||\mathbf{z}^{vMF}_{i}||=1.
\label{eq::recompose}
\end{equation}
After obtaining $\mathbf{\widetilde{Z}}$ as the approximation of $\mathbf{Z}$, the reconstruction network $\bm{R}_\omega$ reconstructs the input image with $\mathbf{\widetilde{Z}}$ as the input, i.e. $\mathbf{\hat{X}}=\bm{R}_\omega(\mathbf{\widetilde{Z}})$.

\subsection{Performing downstream task}
As the vMF likelihoods contain only spatial information of the image that is highly correlated to the segmentation mask, we design a segmentation module, i.e. the task network $\bm{T}_\theta$, to predict the segmentation mask with the vMF likelihoods as input, i.e. $\mathbf{\hat Y}=\bm{T}_\theta(\mathbf{Z}_{vMF})$. Specifically, the segmentation mask tells what anatomical part the feature vector $\mathbf{z}_i$ corresponds to, which provides further guidance for the model to learn the vMF kernels as the components of the anatomical parts. Then the vMF likelihoods will be further aligned when trained with multi-domain data and hence perform well on domain generalisation tasks.

\subsection{Learning objective}
Overall, the model contains trainable parameters $\psi$, $\theta$, $\omega$ and the kernels $\boldsymbol \mu$. The model can be trained end-to-end with the following objective:
\begin{equation}
    \operatorname*{argmin}_{\psi, \theta, \omega, \boldsymbol \mu} \lambda_{Dice} \mathcal{L}_{Dice}(\mathbf{Y}, \hat{\mathbf{Y}}) + \mathcal{L}_{rec}(\mathbf{X}, \hat{\mathbf{X}}) + \mathcal{L}_{vMF}(\boldsymbol \mu, \mathbf{Z}),
    \label{Eqa::totalloss}
\end{equation}
where $\lambda_{Dice}=1$ when the ground-truth mask $\mathbf{Y}$ is available, otherwise $\lambda_{Dice}=0$. $\mathcal{L}_{Dice}$ is the Dice loss \cite{milletari2016v}, $\mathcal{L}_{rec}$ is the reconstruction loss (we use $L1$ distance).

\section{Experiments}
\subsection{Datasets and baseline models}
To make comparison easy, we adopt the datasets and baselines of \cite{liu2021semi}, which we briefly summarise below. We also include \cite{liu2021semi} as a baseline.

\noindent \textbf{Datasets}: The \textbf{Multi-centre, multi-vendor \& multi-disease cardiac image segmentation (M$\&$Ms) dataset \cite{mnms}} consists of 320 subjects scanned at 6 clinical centres using 4 different magnetic resonance scanner vendors i.e. domains A, B, C and D. For each subject, only the end-systole and end-diastole phases are annotated. Voxel resolutions range from $0.85\times 0.85\times 10$ mm to $1.45\times 1.45\times 9.9$ mm. Domain A contains 95 subjects, domain B contains 125 subjects, and domains C and D contain 50 subjects each. The \textbf{Spinal cord gray matter segmentation (SCGM) dataset \cite{prados2017spinal}} images are collected from 4 different medical centres with different MRI systems i.e. domains 1, 2, 3 and 4. The voxel resolutions range from $0.25\times0.25\times2.5$ mm to $0.5\times0.5\times5$ mm. Each domain has 10 labeled subjects and 10 unlabelled subjects.

\noindent \textbf{Baselines}: For fair comparison, we compare all models with the same backbone feature extractor, i.e.\ UNet \cite{ronneberger2015u}, without any pre-training. \textbf{nnUNet \cite{isensee2021nnu}} is a supervised baseline. It adapts its model design and searches the optimal hyperparameters to achieve the optimal performance. \textbf{SDNet+Aug. \cite{liu2020disentangled}} is a semi-supervised disentanglement model, which disentangles the input image to a spatial anatomy and a
non-spatial modality factors. Augmenting the training data by mixing the anatomy and modality factors of different source domains, ``SDNet+Aug." can potentially generalise to unseen domains. \textbf{LDDG \cite{li2020domain}} is a fully-supervised domain generalisation model, in which low-rank regularisation is used and the features are aligned to Gaussian distributions. \textbf{SAML \cite{liu2020shape}} is a gradient-based meta-learning approach. It applies the compactness and smoothness constraints to learn domain-invariant features across meta-train and meta-test sets in a fully supervised setting. \textbf{DGNet \cite{liu2021semi}} is a semi-supervised gradient-based meta-learning approach. Combining meta-learning and disentanglement, the shifts between domains are captured in the disentangled representations. DGNet achieved the state-of-the-art (SOTA) domain generalisation performance on M$\&$Ms and SCGM datasets.

\subsection{Implementation details}
All models are trained using the Adam optimiser \cite{kingma2014adam} with a learning rate of $1\times e^{-4}$ for 50K iterations using batch size 4. Images are cropped to $288\times 288$ for M\&Ms and $144\times 144$ for SCGM. $\bm{F}_\psi$ is a 2D UNet \cite{ronneberger2015u} without the last upsampling and output layers to extract features $\mathbf{Z}$. Note that $\bm{F}_\psi$ can be easily replaced by other encoders such as a ResNet \cite{he2016deep} and the feature vectors can be extracted from any layer of the encoder where performance may vary for different layers. $\bm{T}_\theta$ and $\bm{R}_\omega$ are two shallow convolutional networks that are detailed in Section 2 of Appendix. We follow \cite{kortylewski2020compositional} to set the variance of the vMF distributions as 30. The number of kernels is set to 12, as this number performed the best according to early experiments. For different medical datasets, the best number of kernels may be slightly different. All models are implemented in PyTorch \cite{paszke2019pytorch} and are trained using an NVIDIA 2080 Ti GPU. In the semi-supervised setting, we use specific percentages of the subjects as labeled data and the rest as unlabeled data. We train the models with 3 source domains and treat the 4$^{th}$ domain as the target one. We use Dice (\%) and Hausdorff Distance (HD) \cite{dubuisson1994modified} as the evaluation metrics.

\begin{table}[t]
\centering
\caption{Average Dice (\%) and Hausdorff Distance (HD) results and the standard deviations on M\&Ms and SCGM datasets. For semi-supervised approaches, the training data contain all unlabeled data and different percentages of labeled data from source domains. The rest are trained with different percentages of labeled data only. Results of baseline models are taken from \cite{liu2021semi}. Bold numbers denote the best performance. }\label{tab1}
\begin{tabular}{|c|c|c|c|c|c|c|c|}
\hline
\textbf{Percent} & \textbf{metrics}\         & \textbf{nnUNet}   & \textbf{SDNet+Aug.}  & \textbf{LDDG}    & \textbf{SAML}  & \textbf{DGNet} & \textbf{vMFNet} \\ \cline{1-8}
\multirow{2}{*}{M\&Ms 2\%}     & Dice ($\uparrow$)      & $65.94_{ 8.3}$             & $68.28_{ 8.6}$     & $63.16_{ 5.4}$             & $64.57_{ 8.5}$    & $72.85_{ 4.3}$ & $\mathbf{78.43_{ 3.6}}$ \\ \cline{2-8}
     & HD ($\downarrow$)     & $20.96_{ 4.0}$             & $20.17_{ 3.3}$     & $22.02_{ 3.5}$             & $21.22_{ 4.1}$    & $19.32_{ 2.8}$ & $\mathbf{16.56_{ 1.7}}$ \\ \cline{1-8}
\hline
\multirow{2}{*}{M\&Ms 5\%}     & Dice ($\uparrow$)      & $76.09_{ 6.3}$             & $77.47_{ 3.9}$    & $71.29_{ 3.6}$            & $74.88_{ 4.6}$   & $79.75_{ 4.4}$ & $\mathbf{82.12_{ 3.1}}$ \\ \cline{2-8}
                         & HD ($\downarrow$)      & $18.22_{ 3.0}$             & $18.62_{ 3.1}$    & $19.21_{ 3.0}$            & $18.49_{ 2.9}$   & $17.98_{ 3.2}$  & $\mathbf{15.30_{ 1.8}}$ \\ \cline{1-8}
\multirow{2}{*}{M\&Ms 100\%}     & Dice ($\uparrow$)      & $84.87_{ 2.5}$            & $84.29_{ 1.6}$    & $85.38_{ 1.6}$            & $83.49_{ 1.3}$   & $\mathbf{86.03_{ 1.7}}$ & $85.92_{ 2.0}$ \\ \cline{2-8}
                         & HD ($\downarrow$)      & $14.80_{ 1.9}$            & $15.06_{ 1.6}$    & $14.88_{ 1.7}$            & $15.52_{ 1.5}$   & $14.53_{ 1.8}$ & $\mathbf{14.05_{ 1.3}}$ \\ \cline{1-8}
\hline
\multirow{2}{*}{SCGM 20\%}     & Dice ($\uparrow$)       & $64.85_{ 5.2}$    & $76.73_{ 11}$           & $63.31_{ 17}$   & $73.50_{ 12}$               & $79.58_{ 11}$ & $\mathbf{81.11_{ 8.8}}$ \\ \cline{2-8}
     & HD ($\downarrow$)      & $3.49_{ 0.49}$    & $2.07_{ 0.36}$           & $2.38_{ 0.39}$   & $2.11_{ 0.37}$               & $1.97_{ 0.30}$  & $\mathbf{1.96_{ 0.31}}$ \\ \cline{1-8}
\hline
\multirow{2}{*}{SCGM 100\%}     & Dice ($\uparrow$)       & $71.51_{ 5.4}$    & $81.37_{ 11}$           & $79.29_{ 13}$   & $80.95_{ 13}$              & $82.25_{ 11}$ & $\mathbf{84.03_{ 8.0}}$ \\ \cline{2-8}
                         & HD ($\downarrow$)      & $3.53_{ 0.45}$    & $1.93_{ 0.36}$           & $2.11_{ 0.41}$   & $1.95_{ 0.38}$              & $1.92_{ 0.31}$ & $\mathbf{1.84_{ 0.31}}$ \\ \cline{1-8}
\hline
\end{tabular}
\end{table}

\subsection{Semi-supervised domain generalisation}
\label{sec::results}
Table \ref{tab1} reports the average results over four leave-one-out experiments that treat each domain in turn as the target domain; more detailed results can be found in Section 1 of the Appendix. We highlight that the proposed vMFNet is \textbf{14 times faster to train} compared to the previous SOTA DGNet. Training vMFNet for one epoch takes 7 minutes, while DGNet needs 100 minutes for the M\&Ms dataset due to the need to construct new computational graphs for the meta-test step in every iteration.

%\noindent \textbf{Results}: 
With limited annotations, vMFNet achieves 7.7\% and 3.0\% improvements (in Dice) for 2\% and 5\% cases compared to the previous SOTA DGNet on M\&Ms dataset. For the 100\% case, vMFNet and DGNet have similar performance of around 86\% Dice and 14 HD. Overall, vMFNet has consistently better performance for almost all scenarios on the M\&Ms dataset. Similar improvements are observed for the SCGM dataset.

\noindent \textbf{Which losses help more?} We ablate two key losses of vMFNet in the 2\% of M\&Ms setting. Note that both losses do not require the ground-truth masks. Removing $\mathcal{L}_{rec}$ results in 74.83\% Dice and 18.57 HD, whereas removing $\mathcal{L}_{vMF}$ gives 75.45\% Dice and 17.53 HD. Removing both gives 74.70\% Dice and 18.25 HD. Compared with 78.43\% Dice and 16.56 HD, training with both losses gives better generalisation results when the model is trained to learn better kernels and with unlabeled data. When removing the two losses, the model can still perform adequately compared to the baselines due to the decomposing mechanism.

\noindent \textbf{Alignment analysis}: To show that the vMF likelihoods from different source domains are aligned, for M\&Ms 100\% cases, we first mask out the non-heart part of the images, features and vMF likelihoods. Then, we train classifiers to predict which domain the input is from with the masked images $\mathbf{X}$ or masked features $\mathbf{Z}$ or masked vMF likelihoods $\mathbf{Z}_{vMF}$ as input. The average cross-entropy errors are 0.718, 0.701 and 0.756, which means that it is harder to tell the domain class with the heart part of $\mathbf{Z}_{vMF}$, i.e. the vMF likelihoods for the downstream task are better aligned compared to the features $\mathbf{Z}$ from different source domains.

\subsection{Visualisation of compositionality}
\label{sec::explain}
\begin{figure}[t]
\includegraphics[width=\textwidth]{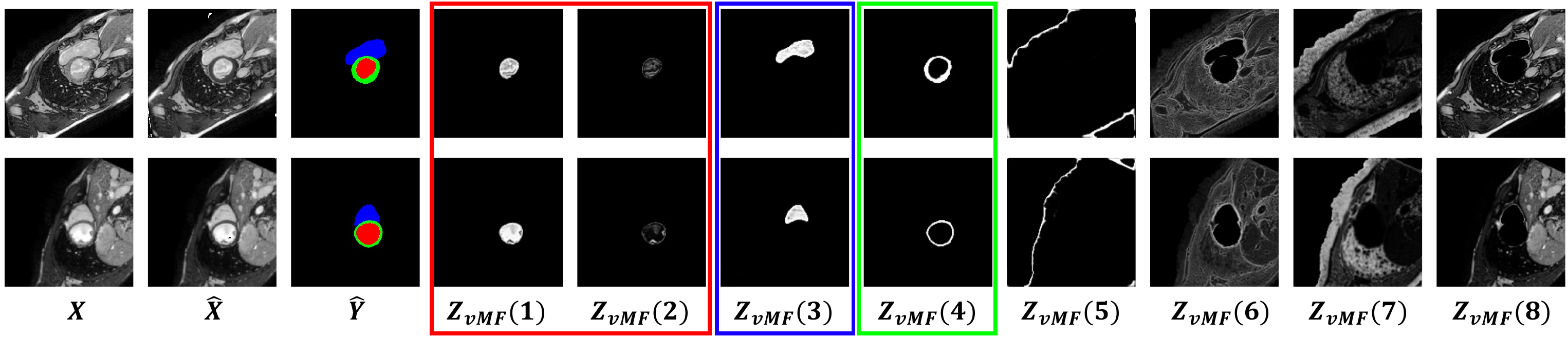}
\centering
\caption{Visualisation of images, reconstructions, predicted segmentation masks and 8 (of 12) most informative vMF likelihood channels for 2 examples from M\&Ms dataset.} \label{fig::visuals}
\end{figure}

Overall, the segmentation prediction can be interpreted as the activation of corresponding kernels at each position, where false predictions occur when the wrong kernels are activated i.e. the wrong vMF likelihoods (composing operations) are used to predict the mask. We show example images, reconstructions, predicted segmentation masks, and 8 most informative vMF likelihoods channels in Fig. \ref{fig::visuals}. As shown, vMF kernels 1 and 2 (red box) are mostly activated by the left ventricle (LV) feature vectors and kernels 3 (blue box) and 4 (green box) are mostly for right ventricle (RV) and myocardium (MYO) feature vectors. Interestingly, kernel 2 is mostly activated by papillary muscles in the left ventricle even though no supervision about the papillary muscles is provided during training. This supports that the model learns the kernels as the compositional components (patterns of papillary muscles, LV, RV and MYO) of the heart. 

\subsection{Test-time domain generalisation}
As discussed in Section \ref{sec::explain}, poor segmentation predictions are usually caused by the wrong kernels being activated. This results in the wrong vMF likelihoods being used to predict masks. Reconstruction quality is also affected by wrong vMF likelihoods. In fact, average reconstruction error is approximately 0.007 on the training set and 0.011 on the test set. Inspired by \cite{he2021autoencoder, valvano2021re} we perform test-time training (TTT) to better reconstruct by fine-tuning the reconstruction loss $\mathcal{L}_{rec}(\mathbf{X}, \hat{\mathbf{X}})$ to update $\bm{F}_\psi$ and $\bm{R}_\omega$ with the kernels and $\bm{T}_\theta$ fixed. This should in turn produce better vMF likelihoods. For images of each subject at test time, we fine-tune the reconstruction loss for 15 iterations (saving the model at each iteration) with a small learning rate of $1\times e^{-6}$. Out of the 15 models, we choose the one with minimum reconstruction error to predict the segmentation masks for each subject. The detailed results of TTT for M\&Ms are included in Section 1 of Appendix. For M\&Ms 2\%, 5\% and 100\% cases, TTT gives around 3.5\%, 1.4\% and 1\% improvements in Dice compared to results (without TTT) in Table~\ref{tab1}.

\section{Conclusion}
In this paper, inspired by the human recognition process, we have proposed a semi-supervised compositional model, vMFNet, that is effective for domain-generalised medical image segmentation. With extensive experiments, we showcased that vMFNet achieves consistently improved performance in terms of semi-supervised domain generalisation and highlighted the correlation between downstream segmentation task performance and the reconstruction. Based on the presented results, we are confident that freezing the kernels and fine-tuning the reconstruction to learn better composing operations (vMF likelihoods) at test time gives further improved generalisation performance.

\section{Acknowledgement}
This work was supported by the University of Edinburgh, the Royal Academy of Engineering and Canon Medical Research Europe by a PhD studentship to Xiao Liu. This work was partially supported by the Alan Turing Institute under the EPSRC grant EP/N510129/1. S.A. Tsaftaris acknowledges the support of Canon Medical and the Royal Academy of Engineering and the Research Chairs and Senior Research Fellowships scheme (grant RCSRF1819\textbackslash8\textbackslash25).

\bibliographystyle{splncs04}
\bibliography{references}

\begin{thebibliography}{10}
\providecommand{\url}[1]{\texttt{#1}}
\providecommand{\urlprefix}{URL }
\providecommand{\doi}[1]{https://doi.org/#1}

\bibitem{achille2018emergence}
Achille, A., Soatto, S.: Emergence of invariance and disentanglement in deep
  representations. JMLR  \textbf{19}(1),  1947--1980 (2018)

\bibitem{arad2021compositional}
Arad~Hudson, D., Zitnick, L.: Compositional transformers for scene generation.
  In: NeurIPS (2021)

\bibitem{bernard2018deep}
Bernard, O., Lalande, A., Zotti, C., Cervenansky, F., Yang, X., et~al.: Deep
  learning techniques for automatic mri cardiac multi-structures segmentation
  and diagnosis: is the problem solved? IEEE TMI  \textbf{37}(11),  2514--2525
  (2018)

\bibitem{mnms}
Campello, V.M., Gkontra, P., Izquierdo, C., Mart{\'\i}n-Isla, C., Sojoudi, A.,
  Full, P.M., Maier-Hein, K., Zhang, Y., He, Z., Ma, J., et~al.: Multi-centre,
  multi-vendor and multi-disease cardiac segmentation: The {M\&Ms} challenge.
  IEEE TMI  (2021)

\bibitem{carlucci2019domain}
Carlucci, F.M., D'Innocente, A., Bucci, S., Caputo, B., Tommasi, T.: Domain
  generalisation by solving jigsaw puzzles. In: CVPR. pp. 2229--2238 (2019)

\bibitem{chartsias2019disentangled}
Chartsias, A., Joyce, T., et~al.: Disentangled representation learning in
  cardiac image analysis. MedIA  \textbf{58},  101535 (2019)

\bibitem{chen2021cooperative}
Chen, C., Hammernik, K., Ouyang, C., Qin, C., Bai, W., Rueckert, D.:
  Cooperative training and latent space data augmentation for robust medical
  image segmentation. In: MICCAI. pp. 149--159. Springer (2021)

\bibitem{chen2020frontiers}
Chen, C., Qin, C., Qiu, H., et~al.: Deep learning for cardiac image
  segmentation: {A} review. Frontiers in Cardiovascular Medicine
  \textbf{7}(25),  1--33 (2020)

\bibitem{dou2019domain}
Dou, Q., Castro, D.C., Kamnitsas, K., Glocker, B.: Domain generalisation via
  model-agnostic learning of semantic features. In: NeurIPS (2019)

\bibitem{dubuisson1994modified}
Dubuisson, M.P., Jain, A.K.: A modified hausdorff distance for object matching.
  In: ICPR. vol.~1, pp. 566--568. IEEE (1994)

\bibitem{gu2021domain}
Gu, R., Zhang, J., Huang, R., Lei, W., Wang, G., Zhang, S.: Domain composition
  and attention for unseen-domain generalizable medical image segmentation. In:
  MICCAI. pp. 241--250. Springer (2021)

\bibitem{he2016deep}
He, K., Zhang, X., Ren, S., Sun, J.: Deep residual learning for image
  recognition. In: CVPR. pp. 770--778 (2016)

\bibitem{he2021autoencoder}
He, Y., Carass, A., Zuo, L., et~al.: Autoencoder based self-supervised
  test-time adaptation for medical image analysis. MedIA  \textbf{72},  102136
  (2021)

\bibitem{hu2021fully}
Hu, M., Song, T., Gu, Y., Luo, X., et~al.: Fully test-time adaptation for image
  segmentation. In: MICCAI. pp. 251--260. Springer (2021)

\bibitem{huang2021fsdr}
Huang, J., Guan, D., Xiao, A., Lu, S.: {FSDR}: Frequency space domain
  randomization for domain generalization. In: CVPR (2021)

\bibitem{huynh2020compositional}
Huynh, D., Elhamifar, E.: Compositional zero-shot learning via fine-grained
  dense feature composition. In: NeurIPS. vol.~33, pp. 19849--19860 (2020)

\bibitem{isensee2021nnu}
Isensee, F., Jaeger, P.F., et~al.: nn{UNet}: a self-configuring method for deep
  learning-based biomedical image segmentation. Nature Methods  \textbf{18}(2),
   203--211 (2021)

\bibitem{iwasawa2021test}
Iwasawa, Y., Matsuo, Y.: Test-time classifier adjustment module for
  model-agnostic domain generalization. In: NeurIPS. vol.~34 (2021)

\bibitem{karani2021test}
Karani, N., Erdil, E., Chaitanya, K., Konukoglu, E.: Test-time adaptable neural
  networks for robust medical image segmentation. MedIA  \textbf{68},  101907
  (2021)

\bibitem{kingma2014adam}
Kingma, D.P., Ba, J.: Adam: {A} method for stochastic optimization. In: ICLR
  (2015)

\bibitem{kortylewski2020compositional}
Kortylewski, A., He, J., Liu, Q., Yuille, A.L.: Compositional convolutional
  neural networks: {A} deep architecture with innate robustness to partial
  occlusion. In: CVPR. pp. 8940--8949 (2020)

\bibitem{li2018learning}
Li, D., Yang, Y., Song, Y.Z., Hospedales, T.: Learning to generalise:
  Meta-learning for domain generalisation. In: AAAI (2018)

\bibitem{li2020domain}
Li, H., Wang, Y., Wan, R., Wang, S., et~al.: Domain generalisation for medical
  imaging classification with linear-dependency regularization. In: NeurIPS
  (2020)

\bibitem{li2021atrialgeneral}
Li, L., Zimmer, V.A., et~al.: Atrialgeneral: Domain generalization for left
  atrial segmentation of {M}ulti-center {LGE} {MRI}s. In: MICCAI. pp. 557--566.
  Springer (2021)

\bibitem{liu2021learning}
Liu, N., Li, S., Du, Y., Tenenbaum, J., Torralba, A.: Learning to compose
  visual relations. In: NeurIPS. vol.~34 (2021)

\bibitem{liu2021feddg}
Liu, Q., Chen, C., Qin, J., Dou, Q., Heng, P.A.: Feddg: Federated domain
  generalization on medical image segmentation via episodic learning in
  continuous frequency space. In: CVPR. pp. 1013--1023 (2021)

\bibitem{liu2020shape}
Liu, Q., Dou, Q., Heng, P.A.: Shape-aware meta-learning for generalising
  prostate mri segmentation to unseen domains. In: MICCAI. pp. 475--485.
  Springer (2020)

\bibitem{liu2020disentangled}
Liu, X., Thermos, S., Chartsias, A., O'Neil, A., Tsaftaris, S.A.: Disentangled
  representations for domain-generalised cardiac segmentation. In: STACOM
  Workshop (2020)

\bibitem{liu2021semi}
Liu, X., Thermos, S., O’Neil, A., Tsaftaris, S.A.: Semi-supervised
  meta-learning with disentanglement for domain-generalised medical image
  segmentation. In: MICCAI. pp. 307--317. Springer (2021)

\bibitem{liu2020measuring}
Liu, X., Thermos, S., Valvano, G., Chartsias, A., O'Neil, A., Tsaftaris, S.A.:
  Measuring the biases and effectiveness of content-style disentanglement. In:
  BMVC (2021)

\bibitem{milletari2016v}
Milletari, F., Navab, N., Ahmadi, S.A.: {VNet}: Fully convolutional neural
  networks for volumetric medical image segmentation. In: 3DV. pp. 565--571.
  IEEE (2016)

\bibitem{paszke2019pytorch}
Paszke, A., Gross, S., Massa, F., Lerer, A., et. al: {PyTorch}: {A}n imperative
  style, high-performance deep learning library. In: NeurIPS. pp. 8026--8037
  (2019)

\bibitem{prados2017spinal}
Prados, F., Ashburner, J., Blaiotta, C., Brosch, T., et~al.: Spinal cord grey
  matter segmentation challenge. Neuroimage  \textbf{152},  312--329 (2017)

\bibitem{ronneberger2015u}
Ronneberger, O., Fischer, P., Brox, T.: {UNet}: Convolutional networks for
  biomedical image segmentation. In: MICCAI. pp. 234--241. Springer (2015)

\bibitem{snell2017prototypical}
Snell, J., Swersky, K., Zemel, R.: Prototypical networks for few-shot learning.
  In: NeurIPS. pp. 4080--4090 (2017)

\bibitem{tokmakov2019learning}
Tokmakov, P., Wang, Y.X., Hebert, M.: Learning compositional representations
  for few-shot recognition. In: CVPR. pp. 6372--6381 (2019)

\bibitem{valvano2021re}
Valvano, G., Leo, A., Tsaftaris, S.A.: Re-using adversarial mask discriminators
  for test-time training under distribution shifts. arXiv preprint
  arXiv:2108.11926  (2021)

\bibitem{valvano2021stop}
Valvano, G., Leo, A., Tsaftaris, S.A.: Stop throwing away discriminators!
  re-using adversaries for test-time training. In: DART Workshop, pp. 68--78.
  Springer (2021)

\bibitem{yao2022enhancing}
Yao, H., Hu, X., Li, X.: Enhancing pseudo label quality for
  semi-superviseddomain-generalized medical image segmentation. arXiv preprint
  arXiv:2201.08657  (2022)

\bibitem{yuan2021robust}
Yuan, X., Kortylewski, A., et~al.: Robust instance segmentation through
  reasoning about multi-object occlusion. In: CVPR. pp. 11141--11150 (2021)

\bibitem{zakazov2021anatomy}
Zakazov, I., Shirokikh, B., et~al.: Anatomy of domain shift impact on {U-Net}
  layers in {MRI} segmentation. In: MICCAI. pp. 211--220. Springer (2021)

\bibitem{zhang2020generalising}
Zhang, L., Wang, X., Yang, D., Sanford, T., et~al.: Generalising deep learning
  for medical image segmentation to unseen domains via deep stacked
  transformation. IEEE TMI  \textbf{39}(7),  2531--2540 (2020)

\bibitem{zhang2021light}
Zhang, Y., Kortylewski, A., Liu, Q., et~al.: A light-weight interpretable
  compositionalnetwork for nuclei detection and weakly-supervised segmentation.
  arXiv preprint arXiv:2110.13846  (2021)

\end{thebibliography}

\newpage
\section{Appendix}
\subsection{More quantitative results}
We report the results of test-time training on M\&Ms dataset in Table \ref{tabTTT}. In Table \ref{tabA1}, Table \ref{tabA1_hsd}, Table \ref{tabA2} and Table \ref{tabA2_hsd}, we report more detailed Dice (\%) and Hausdorff Distance results of semi-supervised domain generalisation.
\begin{table}[ht]
\setcounter{table}{0}
\renewcommand{\thetable}{A\arabic{table}}
\centering
\caption{Dice (\%) $\vline$ Hausdorff Distance results on M\&Ms dataset with test-time training. Improvements are the comparison between the average results of with or without test-time training.}\label{tabTTT}
\resizebox{0.9\textwidth}{!}{%
\begin{tabular}{|c|c|c|c|c|c|c|}
\hline
\textbf{Percent}     & \textbf{B,C,D$\rightarrow$ A}           & \textbf{A,C,D$\rightarrow$ B}   & \textbf{A,B,D $\rightarrow$ C}  & \textbf{A,B,C$\rightarrow$ D}    & \textbf{Average} & \textbf{Improvement}  \\ \cline{1-7}
2\%         & $77.11$ $\vline$ $18.98$           & $80.23$ $\vline$ $16.82$     & $83.06$ $\vline$ $14.85$             & $84.29$ $\vline$ $14.64$    & $81.17$ $\vline$ $16.32$ & $3.5\%$ $\vline$ $0.14\%$ \\ \cline{1-7}
\hline
5\%         & $78.97$ $\vline$ $17.97$           & $83.83$ $\vline$ $15.18$     & $84.13$ $\vline$ $14.29$             & $86.04$ $\vline$ $13.45$    & $83.24$ $\vline$ $15.22$ & $1.4\%$ $\vline$ $0.05\%$ \\ \cline{1-7}
\hline
100\%         & $83.98$ $\vline$ $15.76$           & $85.75$ $\vline$ $14.28$     & $88.57$ $\vline$ $12.39$             & $88.64$ $\vline$ $12.66$    & $86.74$ $\vline$ $13.77$ & $0.95\%$ $\vline$ $0.2\%$ \\ \cline{1-7} \cline{1-7}
\hline
\end{tabular}}
\end{table}

\begin{table}[b]
\renewcommand{\thetable}{A\arabic{table}}
\centering
\caption{Dice (\%) results and the standard deviations on M\&Ms dataset. Bold numbers denote the best performance.}\label{tabA1}
\resizebox{0.9\textwidth}{!}{%
\begin{tabular}{|c|c|c|c|c|c|c|c|c|}
\hline
\multicolumn{2}{|c|}{\textbf{Source}}     & \textbf{Target}           & \textbf{nnUNet}   & \textbf{SDNet+Aug.}  & \textbf{LDDG}    & \textbf{SAML}  & \textbf{DGNet} & \textbf{vMFNet} \\ \cline{1-9}
\multirow{ 5}{*}{2\%} & B,C,D  & A                  & $52.87_{ 19}$             & $54.48_{ 18}$     & $59.47_{ 12}$             & $56.31_{ 13}$    & $66.01_{ 12}$ & $\mathbf{73.13_{ 9.6}}$ \\ \cline{2-9}

& A,C,D       & B                                   & $64.63_{ 17}$             & $67.81_{ 14}$     & $56.16_{ 14}$             & $56.32_{ 15}$    & $72.72_{ 10}$ & $\mathbf{77.01_{ 7.9}}$ \\ \cline{2-9}

& A,B,D       & C                                   & $72.97_{ 14}$             & $76.46_{ 12}$     & $68.21_{ 11}$             & $75.70_{ 8.7}$   & $77.54_{ 10}$ & $\mathbf{81.57_{ 8.1}}$ \\ \cline{2-9}

& A,B,C       & D                                   & $73.27_{ 11}$             & $74.35_{ 11}$     & $68.56_{ 10}$             & $69.94_{ 9.8}$   & $75.14_{ 8.4}$ & $\mathbf{82.02_{ 6.5}}$ \\ \cline{2-9}
\hline
\multirow{ 5}{*}{5\%} & B,C,D   & A                 & $65.30_{ 17}$             & $71.21_{ 13}$     & $66.22_{ 9.1}$            & $67.11_{ 10}$    & $72.40_{ 12}$ & $\mathbf{77.06_{ 10}}$ \\ \cline{2-9}

& A,C,D       & B                                   & $79.73_{ 10}$             & $77.31_{ 10}$     & $69.49_{ 8.3}$            & $76.35_{ 7.9}$   & $80.30_{ 9.1}$ & $\mathbf{82.29_{ 7.8}}$ \\ \cline{2-9}

& A,B,D       & C                                   & $78.06_{ 11}$             & $81.40_{ 8.0}$    & $73.40_{ 9.8}$            & $77.43_{ 8.3}$   & $82.51_{ 6.6}$ & $\mathbf{84.01_{ 7.3}}$  \\ \cline{2-9}    

& A,B,C       & D                                   & $81.25_{ 8.3}$            & $79.95_{ 7.8}$    & $75.66_{ 8.5}$            & $78.64_{ 5.8}$   & $83.77_{ 5.1}$ & $\mathbf{85.13_{ 6.1}}$ \\ \cline{2-9}
\hline
\multirow{ 5}{*}{100\%} & B,C,D  & A                & $80.84_{ 11}$             & $81.50_{ 7.7}$    & $82.62_{ 6.3}$            & $81.33_{ 7.2}$   & $\mathbf{83.21_{ 7.4}}$ & $82.67_{ 7.2}$ \\ \cline{2-9}

& A,C,D       & B                                   & $\mathbf{86.76_{ 5.8}}$   & $85.04_{ 6.1}$    & $85.68_{ 5.7}$            & $84.15_{ 5.9}$   & $86.53_{ 5.3}$ & $85.95_{ 5.6}$ \\ \cline{2-9}

& A,B,D       & C                                   & $84.92_{ 7.1}$            & $85.64_{ 6.5}$    & $86.49_{ 6.3}$            & $84.52_{ 6.2}$   & $87.22_{ 6.1}$ & $\mathbf{87.80_{ 4.4}}$  \\ \cline{2-9}

& A,B,C       & D                                   & $86.94_{ 5.9}$            & $84.96_{ 5.2}$    & $86.73_{ 6.1}$            & $83.96_{ 5.9}$   & $87.16_{ 4.9}$ & $\mathbf{87.26_{ 4.7}}$  \\ \cline{2-9}
\hline
\end{tabular}}
\end{table}

\begin{table}[ht]
\renewcommand{\thetable}{A\arabic{table}}
\centering
\caption{Hausdorff Distance results and the standard deviations on M\&Ms dataset. Bold numbers denote the best performance.}\label{tabA1_hsd}
\resizebox{0.8\textwidth}{!}{%
\begin{tabular}{|c|c|c|c|c|c|c|c|c|}
\hline
\multicolumn{2}{|c|}{\textbf{Source}}     & \textbf{Target}           & \textbf{nnUNet}   & \textbf{SDNet+Aug.}  & \textbf{LDDG}    & \textbf{SAML}  & \textbf{DGNet} & \textbf{vMFNet} \\ \cline{1-9}
\multirow{ 5}{*}{2\%} & B,C,D  & A                  & $26.48_{ 7.5}$             & $24.69_{ 7.0}$     & $25.56_{ 5.9}$             & $25.57_{ 5.7}$    & $23.55_{ 6.5}$ & $\mathbf{19.14_{ 4.8}}$ \\ \cline{2-9}

& A,C,D       & B                                   & $23.11_{ 6.8}$             & $21.84_{ 6.2}$     & $25.44_{ 5.2}$             & $24.91_{ 5.5}$    & $19.95_{ 6.3}$ & $\mathbf{17.01_{ 3.7}}$ \\ \cline{2-9}

& A,B,D       & C                                   & $16.75_{ 4.6}$             & $16.57_{ 4.2}$     & $18.98_{ 3.9}$             & $16.46_{ 3.5}$   & $16.29_{ 4.0}$ & $\mathbf{15.30_{ 3.5}}$ \\ \cline{2-9}
& A,B,C       & D                                   & $17.51_{ 4.9}$             & $17.57_{ 4.1}$     & $18.08_{ 3.8}$             & $17.94_{ 3.8}$   & $17.48_{ 4.7}$ & $\mathbf{14.80_{ 3.0}}$ \\ \cline{2-9}
\hline
\multirow{ 5}{*}{5\%} & B,C,D   & A                 & $23.04_{ 6.7}$             & $22.84_{ 6.3}$     & $23.35_{ 5.7}$            & $23.10_{ 5.9}$    & $22.55_{ 6.6}$ & $\mathbf{18.19_{ 4.9}}$ \\ \cline{2-9}

& A,C,D       & B                                   & $18.18_{ 4.7}$             & $20.26_{ 5.5}$     & $20.56_{ 4.7}$            & $18.97_{ 4.9}$   & $19.37_{ 6.4}$ & $\mathbf{15.24_{ 3.2}}$ \\ \cline{2-9}

& A,B,D       & C                                   & $16.44_{ 4.2}$             & $16.22_{ 3.9}$    & $17.14_{ 3.3}$            & $16.29_{ 3.2}$   & $15.77_{ 3.8}$ & $\mathbf{14.17_{ 3.3}}$ \\ \cline{2-9}    

& A,B,C       & D                                   & $15.24_{ 4.2}$            & $15.15_{ 3.3}$    & $15.80_{ 3.2}$            & $15.58_{ 3.2}$   & $14.24_{ 2.8}$ & $\mathbf{13.61_{ 2.8}}$ \\ \cline{2-9}
\hline
\multirow{ 5}{*}{100\%} & B,C,D  & A                & $17.86_{ 5.5}$             & $17.39_{ 4.5}$    & $17.48_{ 4.1}$            & $17.70_{ 4.2}$   & $17.28_{ 3.9}$ & $\mathbf{15.99_{ 3.5}}$ \\ \cline{2-9}

& A,C,D       & B                                   & $14.82_{ 3.4}$   & $15.55_{ 3.7}$    & $15.42_{ 3.4}$            & $16.05_{ 3.7}$   & $14.99_{ 3.6}$ & $\mathbf{14.58_{ 3.2}}$ \\ \cline{2-9}

& A,B,D       & C                                   & $13.72_{ 3.3}$            & $13.67_{ 3.0}$    & $13.52_{ 2.8}$            & $14.21_{ 3.3}$   & $13.11_{ 2.8}$ & $\mathbf{12.70_{ 2.8}}$  \\ \cline{2-9}

& A,B,C       & D                                   & $12.81_{ 3.4}$            & $13.64_{ 2.9}$    & $13.11_{ 3.0}$            & $14.12_{ 2.8}$   & $\mathbf{12.72_{ 2.6}}$ & $12.94_{ 2.5}$  \\ \cline{2-9}
\hline
\end{tabular}}
\end{table}

\begin{table}[ht]
\renewcommand{\thetable}{A\arabic{table}}
\centering
\caption{Dice (\%) results and the standard deviations on SCGM dataset. Bold numbers denote the best performance.}\label{tabA2}
\resizebox{0.8\textwidth}{!}{%
\begin{tabular}{|c|c|c|c|c|c|c|c|c|}
\hline
\multicolumn{2}{|c|}{\textbf{Source}}     & \textbf{Target}           & \textbf{nnUNet}               & \textbf{SDNet+Aug.}                & \textbf{LDDG}                & \textbf{SAML}                           & \textbf{DGNet} & \textbf{vMFNet} \\ \cline{1-9}
\multirow{ 5}{*}{20\%} & 2,3,4  & 1                 & $59.07_{ 21}$        & $83.07_{ 16}$               & $77.71_{ 9.1}$      & $78.71_{ 25}$                  & $87.45_{ 6.3}$ & $\mathbf{88.08_{ 6.9}}$ \\ \cline{2-9}

& 1,3,4       & 2                                   & $69.94_{ 12}$        & $80.01_{ 5.2}$              & $44.08_{ 12}$       & $75.58_{ 12}$                  & $81.05_{ 5.2}$ & $\mathbf{81.21_{ 4.2}}$ \\ \cline{2-9}

& 1,2,4       & 3                                   & $60.25_{ 7.2}$       & $58.57_{ 10}$               & $48.04_{ 5.5}$      & $54.36_{ 7.6}$                 & $61.85_{ 7.3}$ & $\mathbf{66.74_{ 4.9}}$ \\ \cline{2-9}    

& 1,2,3       & 4                                   & $70.13_{ 4.3}$       & $85.27_{ 2.2}$              & $83.42_{ 2.7}$      & $85.36_{ 2.8}$                 & $87.96_{ 2.1}$ & $\mathbf{88.39_{ 2.4}}$ \\ \cline{2-9}
\hline
\multirow{ 5}{*}{100\%} & 2,3,4  & 1                & $75.27_{ 8.3}$    & $90.25_{ 4.5}$  & $88.21_{ 4.9}$   & $90.22_{ 5.6}$              & $90.01_{ 4.9}$ & $\mathbf{90.96_{ 4.7}}$ \\ \cline{2-9}

& 1,3,4       & 2                                   & $76.32_{ 2.9}$    & $84.13_{ 4.2}$           & $83.76_{ 3.1}$   & $\mathbf{86.65_{ 3.5}}$     & $85.48_{ 2.3}$ & $84.89_{ 3.2}$ \\ \cline{2-9}

& 1,2,4       & 3                                   & $62.59_{ 6.9}$    & $62.18_{ 10}$            & $56.11_{ 9.3}$   & $58.27_{ 9.4}$              & $64.23_{ 9.7}$  & $\mathbf{70.71_{ 9.2}}$ \\ \cline{2-9}

& 1,2,3       & 4                                   & $71.87_{ 2.5}$    & $88.93_{ 1.9}$           & $89.08_{ 2.7}$   & $88.66_{ 2.6}$              & $89.26_{ 2.5}$  & $\mathbf{89.57_{ 3.1}}$ \\ \cline{2-9}
\hline
\end{tabular}}
\end{table}

\begin{table}[b]
\renewcommand{\thetable}{A\arabic{table}}
\centering
\caption{Hausdorff Distance results and the standard deviations on SCGM dataset. Bold numbers denote the best performance.}\label{tabA2_hsd}
\resizebox{0.8\textwidth}{!}{%
\begin{tabular}{|c|c|c|c|c|c|c|c|c|}
\hline
\multicolumn{2}{|c|}{\textbf{Source}}     & \textbf{Target}           & \textbf{nnUNet}               & \textbf{SDNet+Aug.}                & \textbf{LDDG}                & \textbf{SAML}                           & \textbf{DGNet} & \textbf{vMFNet} \\ \cline{1-9}
\multirow{ 5}{*}{20\%} & 2,3,4  & 1                 & $3.09_{ 0.25}$        & $1.52_{ 0.33}$               & $1.75_{ 0.26}$      & $1.53_{ 0.38}$                  & $1.50_{ 0.30}$ & $\mathbf{1.47_{ 0.33}}$ \\ \cline{2-9}

& 1,3,4       & 2                                   & $3.16_{ 0.09}$        & $1.97_{ 0.16}$              & $2.73_{ 0.33}$       & $2.07_{ 0.35}$                  & $\mathbf{1.91_{ 0.16}}$ & $1.92_{ 0.14}$ \\ \cline{2-9}

& 1,2,4       & 3                                   & $3.38_{ 0.27}$       & $2.45_{ 0.27}$               & $2.67_{ 0.25}$      & $2.52_{ 0.24}$                 & $\mathbf{2.23_{ 0.23}}$ & $2.25_{ 0.16}$ \\ \cline{2-9}    

& 1,2,3       & 4                                   & $4.31_{ 0.14}$       & $2.34_{ 0.21}$              & $2.37_{ 0.14}$      & $2.30_{ 0.18}$                 & $2.22_{ 0.13}$ & $\mathbf{2.18_{ 0.14}}$ \\ \cline{2-9}

\hline
\multirow{ 5}{*}{100\%} & 2,3,4  & 1                & $3.26_{ 0.21}$    & $1.37_{ 0.25}$  & $1.50_{ 0.23}$   & $1.43_{ 0.36}$              & $1.43_{ 0.29}$ & $\mathbf{1.35_{ 0.25}}$ \\ \cline{2-9}

& 1,3,4       & 2                                   & $3.19_{ 0.09}$    & $1.88_{ 0.16}$           & $2.19_{ 0.19}$   & $\mathbf{1.80_{ 0.19}}$     & $1.81_{ 0.15}$ & $\mathbf{1.80_{ 0.19}}$ \\ \cline{2-9}

& 1,2,4       & 3                                   & $3.37_{ 0.27}$    & $2.34_{ 0.24}$            & $2.64_{ 0.28}$   & $2.43_{ 0.33}$              & $2.23_{ 0.32}$ & $\mathbf{2.13_{ 0.30}}$ \\ \cline{2-9}

& 1,2,3       & 4                                   & $4.30_{ 0.15}$    & $2.13_{ 0.17}$           & $2.12_{ 0.15}$   & $2.15_{ 0.15}$              & $2.11_{ 0.13}$ & $\mathbf{2.07_{ 0.18}}$ \\ \cline{2-9}

\hline
\end{tabular}}
\end{table}

\subsection{Implementation details}
$\bm{T}_\theta$ and $\bm{R}_\omega$ have similar structures, where a double CONV layer (kernel size 3, stride size 1 and padding size 1) in UNet with batch normalisation and ReLU is first used to process the features. Then a transposed convolutional layer is used to upsample the features following with a double CONV layer with batch normalisation and ReLU. Finally, a output convolutional layer with $1\times 1$ kernels is used. For $\bm{T}_\theta$, the output of the last layer is processed with Sigmoid. 

\end{document}